\def\year{2019}\relax
\pdfoutput=1
\documentclass[letterpaper]{article} 
\usepackage{aaai}  
\usepackage{times}  
\usepackage{helvet}  
\usepackage{courier}  
\usepackage{url}  
\usepackage{graphicx}  
\usepackage{multirow}
\usepackage{amsfonts}
\usepackage{booktabs}
\usepackage{graphicx}
\usepackage{amsmath}

\usepackage{multirow}
\usepackage{commath}
\usepackage{makecell} 
\usepackage{color}

\newcommand{\ie}{{\it i.e.,} }
\newcommand{\eg} {{\it e.g.,} }

\newcommand{\wrt} {w.r.t. }

\newcommand{\methodname}{GeoGAN}

\title{Instance-level Facial Attributes Transfer with Geometry-Aware Flow}

\author{Weidong Yin \\ University of British Columbia \\ \texttt{wdyin@cs.ubc.ca} \And Ziwei Liu \\ Chinese University of Hong Kong \\\texttt{zwliu@ie.cuhk.edu.hk} \And Chen Change Loy \\ Nanyang Technological University \\ \texttt{ccloy@ntu.edu.sg}}
\begin{document}

\maketitle

\begin{abstract}

We address the problem of instance-level facial attribute transfer without paired training data, \eg faithfully transferring the exact mustache from a source face to a target face.
This is a more challenging task than the conventional semantic-level attribute transfer, which only preserves the generic attribute style instead of instance-level traits.
We propose the use of geometry-aware flow, which serves as a well-suited representation for modeling the transformation between instance-level facial attributes.
Specifically, we leverage the facial landmarks as the geometric guidance to learn the differentiable flows automatically, despite of the large pose gap existed.
Geometry-aware flow is able to warp the source face attribute into the target face context and generate a warp-and-blend result.
To compensate for the potential appearance gap between source and target faces, we propose a hallucination sub-network that produces an appearance residual to further refine the warp-and-blend result.  
Finally, a cycle-consistency framework consisting of both attribute transfer module and attribute removal module is designed, so that abundant unpaired face images can be used as training data.  
Extensive evaluations validate the capability of our approach in transferring instance-level facial attributes faithfully across large pose and appearance gaps.
Thanks to the flow representation, our approach can readily be applied to generate realistic details on high-resolution images\footnote{Project page: \url{http://mmlab.ie.cuhk.edu.hk/projects/attribute-transfer/}}.
 
\end{abstract}

\section{Introduction}



Modeling and manipulating facial attributes have been a long quest in computer vision~\cite{liu2015faceattributes,Nguyen2008ImagebasedS,huang2018deep,loy2017deep}.
On the one hand, facial attributes are one of the most prominent visual traits we perceive in daily life, thus constituting an important visual element to understand.
On the other hand, the ability to manipulate facial attributes can enable lots of useful real-world applications, such as targeted face editing~\cite{Shu2017NeuralFE,Shen2017LearningRI,Brock2016NeuralPE,Yeh2016SemanticFE}. 

\begin{figure}[t]
  \centering
  \includegraphics[height=6cm]{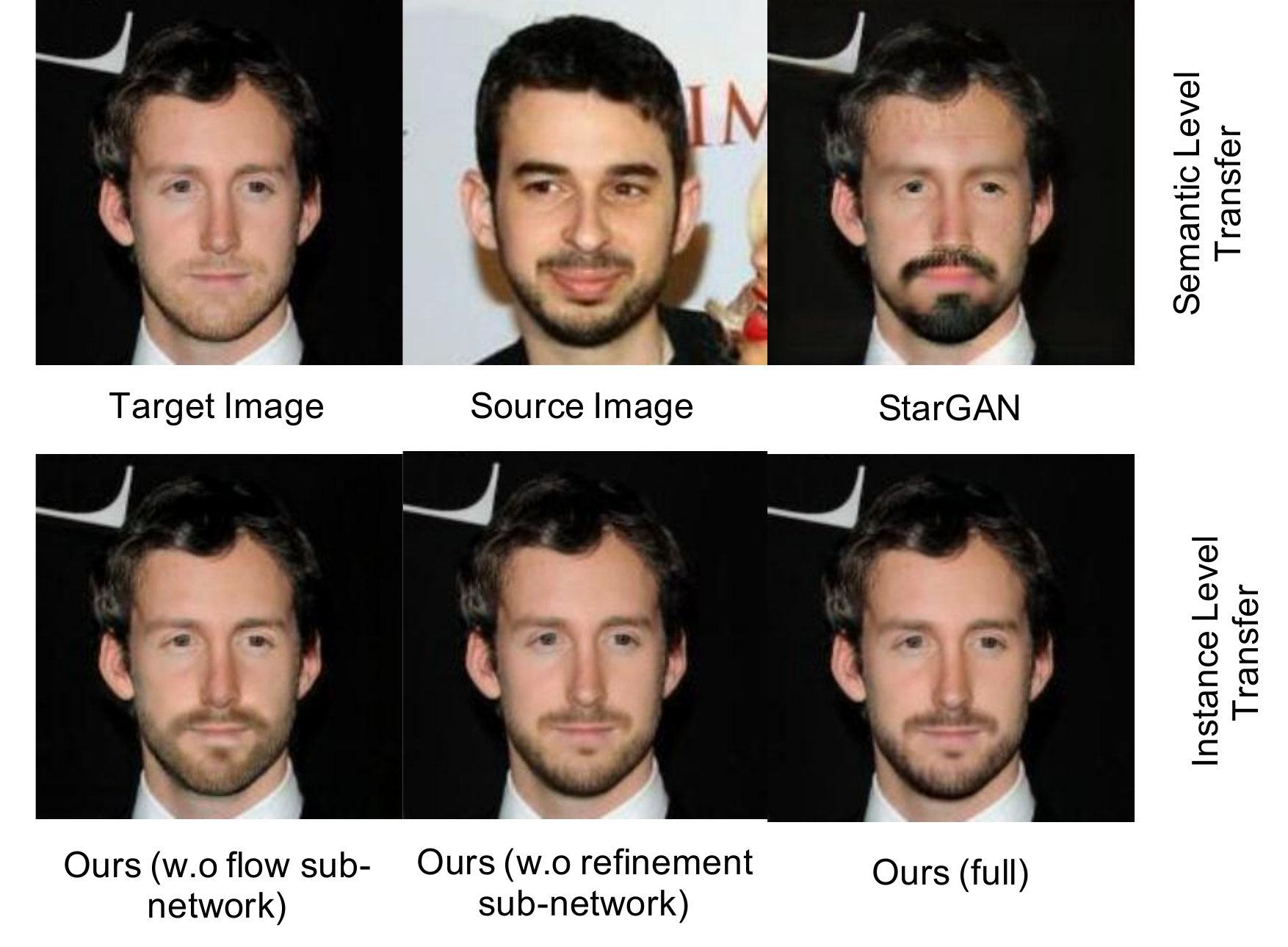}
  \caption{(First row) Semantic-level facial attribute transfer, and (second row) instance-level facial attribute transfer. The former only models the generic attribute style while the latter additionally preserves sample-dependant traits. }
  \label{fig:intro_fig}
\end{figure}

Existing studies on facial attribute manipulation mainly focus on semantic-level attribute transfer~\cite{Perarnau2016InvertibleCG,Lample2017FaderNM,Choi2017StarGANUG,Gardner2015DeepMT}, \ie making the target face possess certain attributes (\eg `mustache') perceptually, as shown in Figure~\ref{fig:intro_fig} (first row). There is, however, no guarantee that the transferred `mustache' on the target face will look alike the one on the source face. 

In this work, we address the problem of instance-level facial attribute transfer without paired training data.
For example, given an unordered collection of images with and without mustache for training, our approach learns to faithfully transfer the exact mustache from the target face to the source face.
This is a more challenging task than the conventional semantic-level attribute transfer. In particular, besides capturing the generic attribute style that are shared by all `mustache' samples, instance-level attribute transfer requires the extraction and preservation of sample-dependant traits, \eg, the mustache from the source image and the lip of the target image, as shown in Figure~\ref{fig:intro_fig} (second row).

\begin{figure*}[t]
  \centering
  \includegraphics[height=6cm]{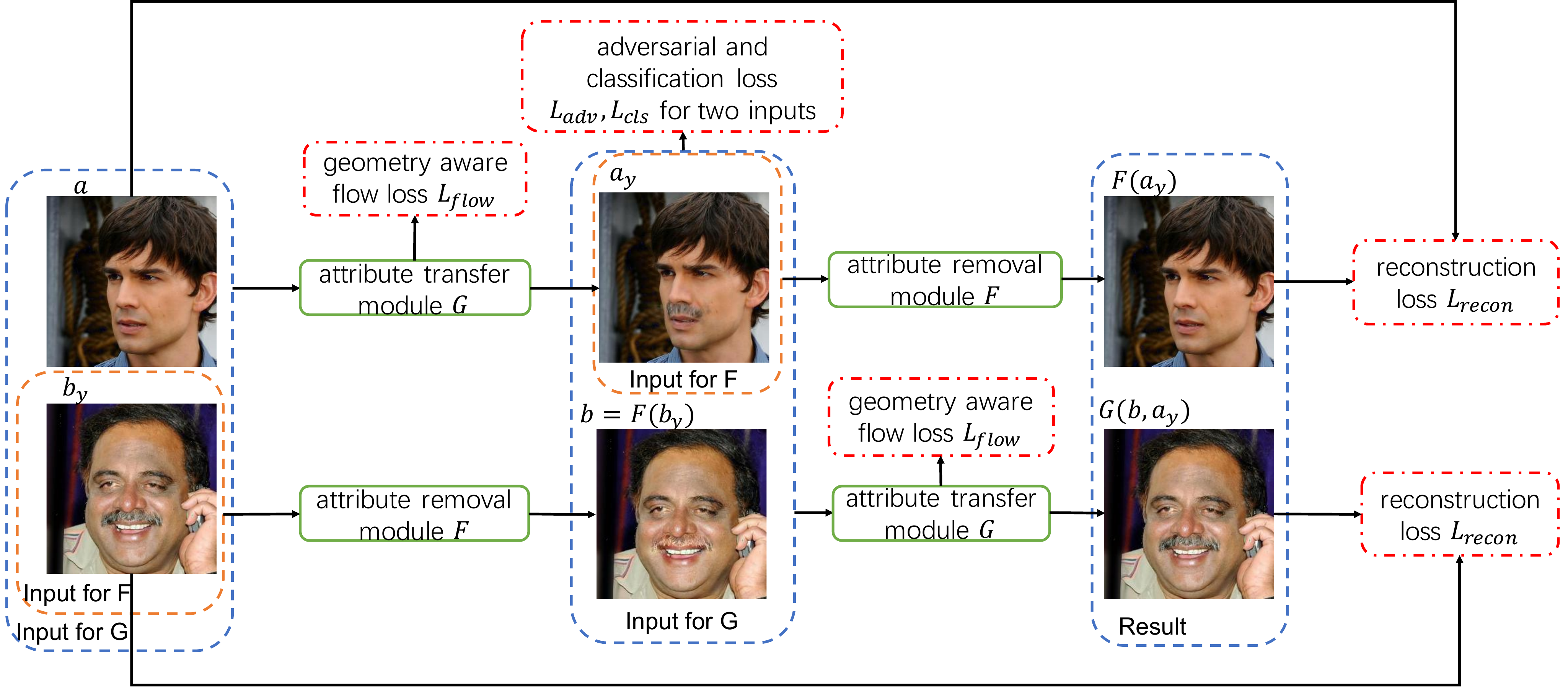}
  \caption{The pipeline of instance-level attribute transfer. Given a target image $a$ and a source image $b_y$, our approach automatically learns an attribute transfer module $G$ and an attribute removal module $F$. These two modules jointly operate in a cycle-consistency manner to learn from abundant unpaired data. 
  }
  \label{fig:overview}
\end{figure*}

The main contribution of our paper is a novel notion of \textbf{\textit{geometry-aware flow}}, which is well-suited as a representation to model the transformation between instance level attributes. 
Unlike the conventional differentiable flow~\cite{Jaderberg2015SpatialTN,Zhou2016ViewSB}, geometric flow is learned under the geometric guidance from facial landmarks. Consequently, the flow can cope robustly with the large displacement gap between facial poses of target and source images. In addition, since flow is invariant to scaling, the learned geometry-aware flow can be readily applied to high-resolution images and generate desired attributes with realistic warp-and-blend details.

To further enhance the transfer quality, we propose a refinement sub-network to rectify potential appearance gaps, \eg, skin color and lighting changes, which cannot be handled well with flow-based warping. This is achieved through generating an appearance residual that can be added to the previous warp-and-blend result. The importance of the refinement network can be observed in Figure~\ref{fig:intro_fig} (second row).  Without the sub-network, the transferred `mustache' on the target face is susceptible to the surrounding skin color of the target face.
Finally, a cycle-consistency framework consisting of an attribute transfer module and an attribute removal module is designed so that learning can be done on abundant unpaired facial images. 
Extensive evaluations on CelebA~\cite{liu2015faceattributes} and CelebA-HQ~\cite{Karras2017ProgressiveGO} datasets validate the effectiveness of our approach in transferring instance-level facial attributes faithfully across large pose and appearance gaps.
 


\section{Related Work}



\noindent
\textbf{Semantic-level Transfer.}
%
Many recent works have achieved impressive results in this direction. 
StarGAN~\cite{Choi2017StarGANUG} applies cycle consistency to preserve identity, and uses classification loss to transfer between different domains. The task of facial attribute transfer can also be viewed as a specific kind of image-to-image translation problem. UNIT~\cite{Liu2017UnsupervisedIT} combines variational autoencoders (VAEs)~\cite{Kingma2013AutoEncodingVB} with CoGAN~\cite{Liu2016CoupledGA} to learn the joint distribution of images in two domains. 
These methods only tackle the task of facial attribute transfer at a semantic level and could not transfer specific attribute of different style at an instance level. 

\noindent
\textbf{Instance-level Transfer.}
There exist some recent studies that discuss the possibility of transferring facial attributes at instance level. GeneGAN~\cite{Zhou2017GeneGANLO} learns to transfer a desired attribute from a reference image by constructing disentangled attribute subspaces from weakly labeled data. 
ELEGANT~\cite{Xiao2018ELEGANTEL} exchanges attribute between two faces by exchanging latent codes of two faces. All these methods are fully parametric, which suffers from blurry results and cannot scale to high-resolution. Some other studies try to directly compose exemplar images. ST-GAN~\cite{Lin2018STGANST} uses a spatial transformer network to transform external objects into the correct position before composing them onto faces. Compositional GAN~\cite{Azadi2018CompositionalGL} learns to model object compositions in a GAN framework. Different from these works, our approach does not require an object to be provided explicitly. Instead, we extract the desired facial attribute automatically from exemplar images to facilitate the faithfulness of instance-level attribute transfer.


\section{Methodology}

\begin{figure}
  \centering
  \includegraphics[width=\linewidth]{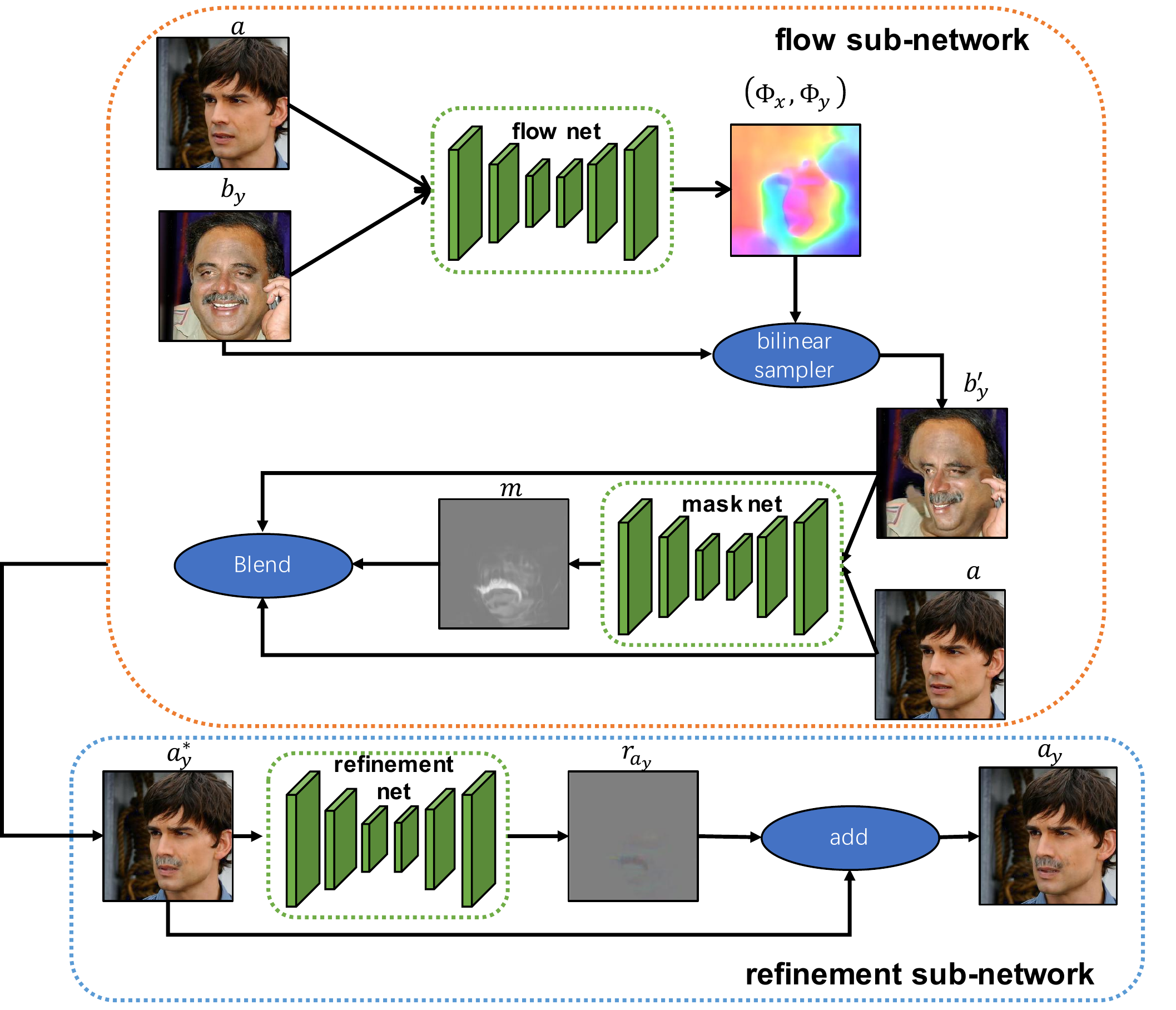}
  \caption{Network architecture of attribute transfer module. Given a source image and a target image, the flow sub-network learns to generate geometric-aware flow to warp the source image into the desired pose. An attention mask is then generated to blend the source and target images. Next, the refinement sub-network synthesizes an appearance residual to compensate for other differences in appearance, like skin colors or lighting conditions.}
  \label{fig:subnetworks}
\end{figure}


%
%
Let $A$ be a domain of target images without certain attribute and $B$ be a domain of source images with the specific kind of attribute we want to transfer, where no pairings exist between the two domains. Given $a\in A$ be a target image without attribute and $b_y \in B$ a source image with the specific kind of attribute we want to transfer, where in the case of people's faces are selected as $eyeglasses,mustache,goatee$. Our goal is to learn a model $G$ that can transfer desired attributes from $b_y$ onto $a$ such that we obtain $a_y$ after the transfer. Different from general attribute manipulation~\cite{Choi2017StarGANUG}, our model transfers attribute at instance level. For the same target image, given different source images we can transfer the same attribute of different styles onto the target image to generate diverse results.

\subsection{Instance-level Attribute Transfer Network}

As shown in Figure~\ref{fig:overview}, two separate modules are introduced to achieve instance-level attribute transfer.

\begin{itemize}
    \item Attribute transfer module $G$. Given an image of a person without some attribute $a$, and an image of a different person with desired attribute $b_y$, the attribute transfer module $G:A \times B \rightarrow B$ extracts the desired attribute from $b_y$ and applies it to $a$ maintaining its identity.  
    \item Attribute removal module $F$.   
    Given the same photo $b_y$, the attribute removal network $F:B \rightarrow A$ learns to remove the attribute while maintaining the identity of $b_y$.
\end{itemize}

Our main focus is on the attribute transfer module. $F$ is an auxiliary part to maintain identity information of $a_y$ during the transfer. Ideally, the output of $G$ as $a_y$ can be used as an example to be transferred to the output of $F$ as $b$, and transferring attributes from $a_y$ to $b$ should generate $b_y$ exactly.

\vspace{0.10cm}
\noindent
\textbf{Attribute Transfer Module.}
The target image $a$ and source image $b_y$ often vary in pose, expression, skin colors and lighting. Taking $b_y$ and $a$ directly into CNNs cannot produce faithful results because some attributes are not aligned well. To address the differences in pose and expression, we introduce a flow sub-network; and for the difference in lighting conditions and skin colors, we devise a refinement sub-network. The networks are shown in Figure~\ref{fig:subnetworks}. These two networks are trained end-to-end. 

\begin{itemize}
\item \textbf{Flow Sub-Network}. The flow sub-network learns to generate the geometry-aware flow and an attention mask. The geometry-aware flow is generated as $\{\Phi^x,\Phi^y\}$, and the source image $b_y$ is warped as $b_y'$. The output pixel of the warped image at location $(i,j)$ is given as
\begin{equation}
\begin{split}
b_y'(i,j) = \sum_{(i',j')\in N} &(1-\abs{i+\Phi^x(i,j)-i'}) \\
&(1-\abs{j+\Phi^y(i,j)-j'})b_y,
\end{split}
\end{equation}
where $N$ stands for 4-pixel neighbors of $(x+\Phi^x,y+\Phi^y)$. We note that $b_y'$ is differentiable to $\{\Phi^x,\Phi^y\}$~\cite{liu2017video}. Thus the flow sub-network can be trained end-to-end and integrated seamlessly into the proposed pipeline.
An attention mask $m$ is learned to select pixels from $a$ and $b_y'$, and the result is blended as $a_y^* = m*a + (1-m) * b_y'$.

\begin{figure}
  \centering
  \includegraphics[width=\linewidth]{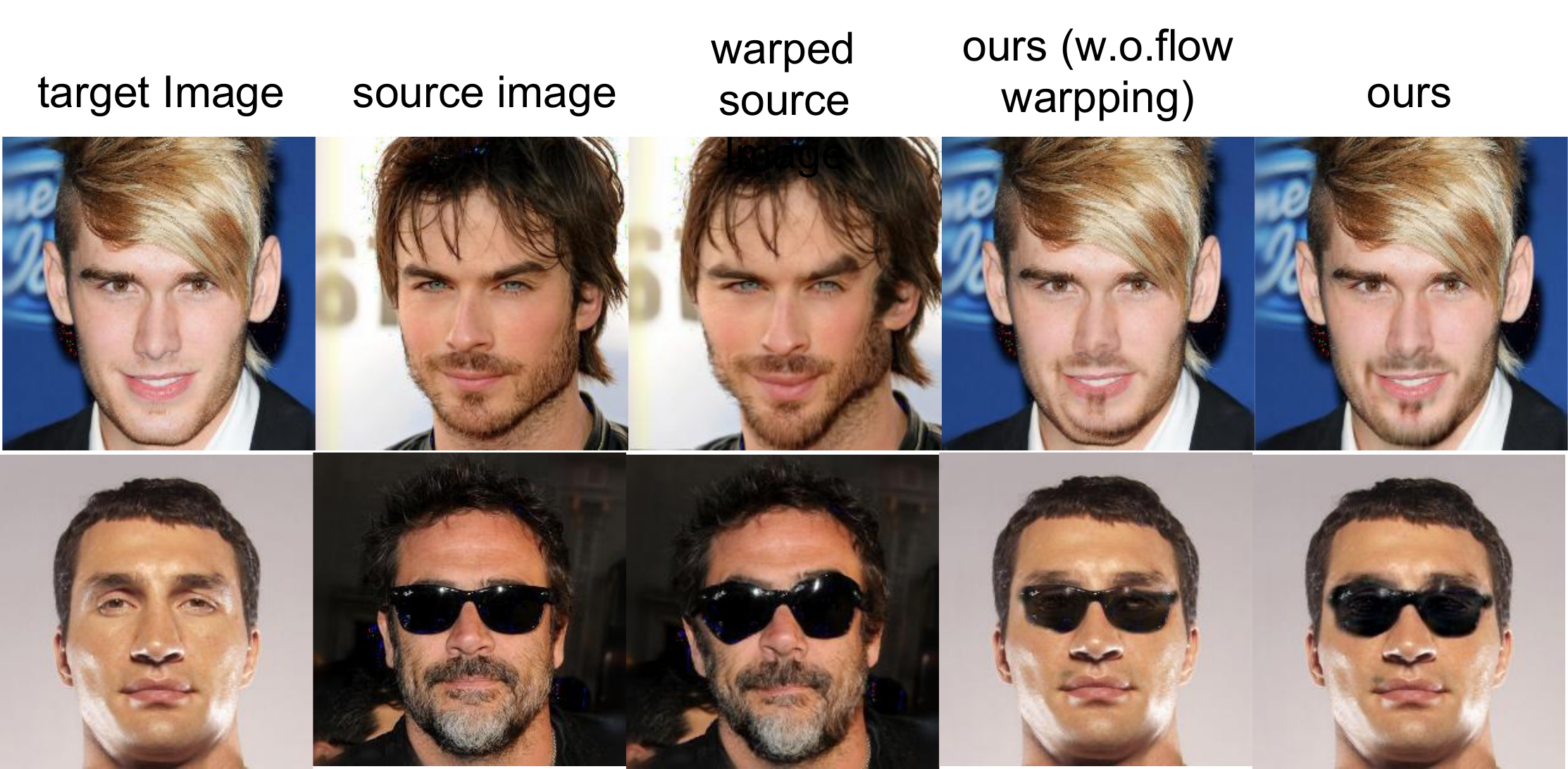}
  \caption{Step-by-step results of geometry-aware flow. The columns are respectively the target image $a$, source image $b_y$, warped source image $b_y'$, result $a_y$ without flow warping and full model's result $a_y$. Without the geometry-aware flow, attributes between the source image and target image are not aligned correctly.}
  \label{fig:flow_ablation}
\end{figure}

The geometry-aware flow is learned to align poses and expressions between $a$ and $b_y$. As shown in Figure~\ref{fig:flow_ablation}, without flow warping the mustache and eyeglasses are misaligned and the method produces erroneous results. More results are shown in Figure~\ref{fig:flowpart_ablation} in the experiments section.

%

\item \textbf{Refinement Sub-Network}. Given $a_y^*$ as input, the refinement sub-network learns to synthesize an appearance residual $r_{a_y}$ to compensate for differences in skin colors and lightings. The final result is generated as $a_y = \alpha r_{a_y} + a_y^*$, where $\alpha$ is a hyper parameter to control the balance between flow sub-network and refinement sub-network. As shown in Figure~\ref{fig:hallucination_ablation}, when skin colors or lighting conditions are different, the sub-network can generate appearance residuals to fix these differences.

\end{itemize}

\begin{figure}[t]
  \centering
  \includegraphics[width=\linewidth]{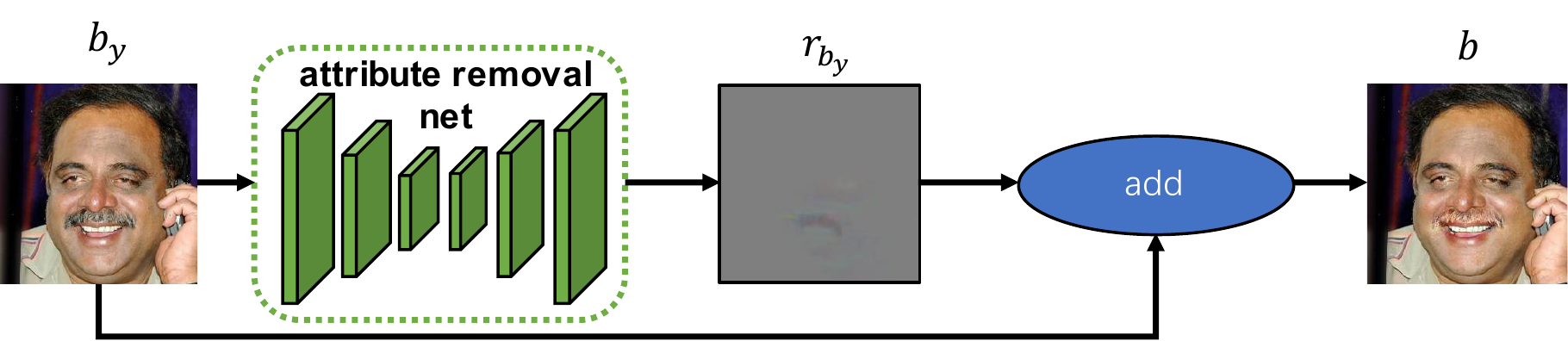}
  \caption{Network architecture of attribute removal module.}
  \label{fig:overview_frn}
\end{figure}

\vspace{0.1cm}
\noindent
\textbf{Attribute Removal Module.}
For the attribute removal module, the input $b_y$ and output $b$ are of the same pose and expression, so we do not need the flow sub-network to warp input image. We adapt an encoder-decoder structure shown in Figure~\ref{fig:overview_frn}. To circumvent information loss, we use an U-Net architecture in which the $i-th$ layer is concatenated to the $(L-i)-th$ layer via skip connections. 
%
%
The network generates an image residual $r_{b_y}$ and adds it onto $b_y$ to get the final image, \ie $b = b_y + r_{b_y}$.

\subsection{Multi-Objective Learning}

Our full learning objective consists of four terms: adversarial, domain classification, reconstruction and geometry-aware flow terms. We detail these terms as follows.


\vspace{0.1cm}
\noindent
\textbf{Adversarial Objective.}
%
To make the images generated by attribute transfer network indistinguishable from real images, we adopt an adversarial loss. To stablize the training process and avoid common training failure modes such as mode collapse and gradient vanishing problem, we replace the original GAN loss as LSGAN~\cite{Mao2017LeastSG} objective: 
\begin{equation}
\begin{split}
L_{adv-g} &= \mathbb{E}_{a\in A, b_y \in B}[\norm{1-D_{src}(G(a,b_y))}^2] \\
		&+  \mathbb{E}_{a\in A}[\norm{D_{src}(b_y)}^2],
\end{split}
\end{equation}
where the discriminator $D$ attempts to discriminate between the real samples and the generated samples, and the attribute transfer network $G$ aims to generate images that cannot be distinguished by the adversary. 

We also apply an adversarial loss to encourage attribute removal network, $F$, to generate images indistinguishable from the faces without certain attribute. 
\begin{equation}
L_{adv-f} = \mathbb{E}_{b_y \in B}[\norm{1-D_{src}(F(b_y))}^2] + \mathbb{E}_{a \in A}[\norm{F(a)}^2].
\end{equation}

\vspace{0.1cm}
\noindent
\textbf{Domain Classification Objective.}
We also introduce an auxiliary classifier to allow our discriminators to constrain the result of attribute transfer and removal to falling within the correct attribute class domains.  We formulate the constrain as a binary classification problem, since we only have two domains: with or without an attribute. The objective can be decomposed into two terms, namely a domain classification loss of real images, which is used to optimize $D$, and a domain classification loss of fake images that is used to optimize $G$. The former is defined as 
\begin{equation}
L_{cls}^r = \mathbb{E}_{a \in A}[\log P(c=0|a)] + \mathbb{E}_{b_y \in B}[\log P(c=1|b_y)]. 
\end{equation}
The latter is 
\begin{equation}
\begin{split}
L_{cls}^f &= \mathbb{E}_{a \in F(b_y)}[\log P(c=0|a)] \\
&+ \mathbb{E}_{a_y\in G(a,b_y)}[\log P(c=1|a_y)] .
\end{split}
\end{equation}

\noindent
\textbf{Reconstruction Objective.}
To ensure that the identity  of a face with transferred attribute is preserved, we introduce and attribute removal network as a constraint. Intuitively, if we apply attribute transfer to $a$ and remove it, then we should get back the original image. Likely, if we remove the attribute on $b_y$ to $a$ as $a_y$ and then transfer back, we should get the same image as $b_y$. 
\begin{equation}
\begin{split}
L_{rec} &= \mathbb{E}_{b_y\in B, a \in A}[\norm{F(G(a,b_y))-a}_1] \\
&+ \mathbb{E}_{a_y\in G(a,b_y),b \in F(b_y)}[\norm{G(b,a_y)-b_y}_1].
\end{split}
\end{equation}

\noindent
\textbf{Geometry-Aware Flow Objective.}
Recall that the generated geometry-aware flow is used to warp a given reference image to align with a target image. Towards this goal, we introduce a landmark loss as well as TV regularization loss to facilitate the training of flow. For landmark loss, we use FAN~\cite{bulat2017far} to detect 68 landmarks $\{x^{b_y}_j,y^{b_y}_j|_{j=1}^{68}\}$ for $b_y$ and $\{x^{a}_j,y^{a}_j|_{j=1}^{68}\}$ for $a$. We require that the landmarks of $b_y$ and $a$ should be close, so the landmark loss is defined as
\begin{equation}
\begin{split}
L_{lm} = \sum\nolimits_{j=1}^{68} &(\Phi^x(x^{b_y}_j,y^{b_y}_j)+x^{b_y}_j - x^a_j)^2 \\
 +&(\Phi^y(x^{b_y}_j,y^{b_y}_j)+y^{b_y}_j-y^a_j)^2.
\end{split}
\end{equation}

As the landmark loss can only be imposed on 68 landmarks, we hope that geometry-aware flow should be spatially smooth so that the structure of target image can be maintained. Thus, a TV regularizer is used. It is defined as 
\begin{equation}
L_{tv} = \norm{\nabla \Phi^x}^2 + \norm{\nabla \Phi^y}^2
\end{equation}

We combine the landmark loss with TV regularizer and define the flow loss as
\begin{equation}
L_{flow} = l_{lm} + l_{tv}.
\end{equation}

\noindent
\textbf{Overall Objective.}
We combine the adversarial, domain classification, reconstruction, and flow objectives to obtain the overall objective as 
\begin{equation}
L_{full} =  L_{adv-f} + L_{adv-g} + L_{cls}^r + L_{cls}^f + L_{rec} + L_{flow}.
\end{equation}
This overall objective is optimized in a multi-task learning manner.
Network hyper-parameters and loss weights are determined on a probe validation subset.

\section{Experiments}


\begin{figure*}[]
  \centering
  \includegraphics[height=8cm]{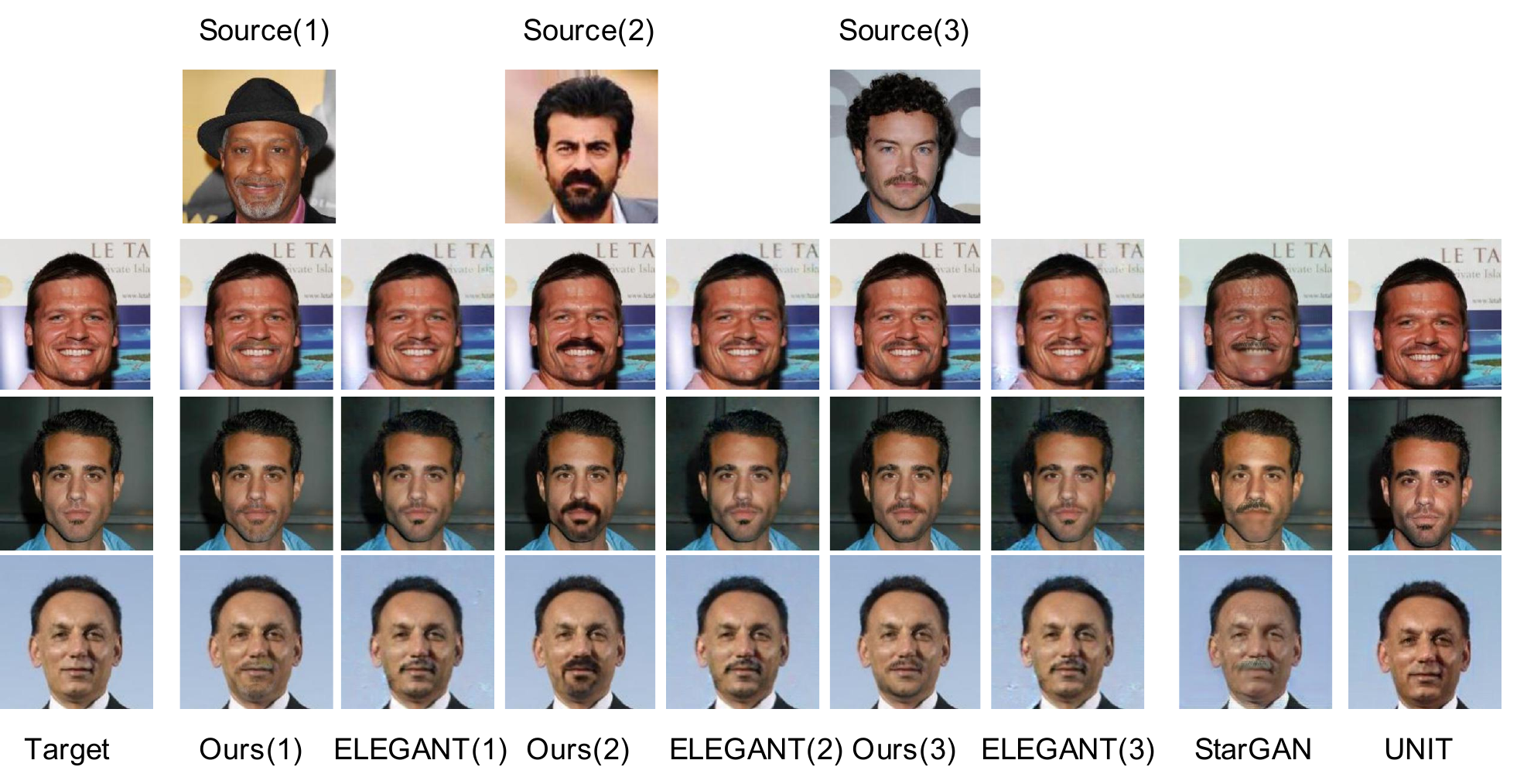}
  \caption{Mustache transfer results on the CelebA dataset. The first six columns the results of the proposed method and ELEGANT~\cite{Xiao2018ELEGANTEL} for three different source images. The last two columns show the result of StarGAN~\cite{Choi2017StarGANUG} and UNIT~\cite{Liu2017UnsupervisedIT}, respectively. 
  } 
  \label{fig:compare_mustache}
\end{figure*}

In this section, we comprehensively evaluate our approach on different benchmarks with dedicated metrics.

\subsection{Implementation Details}

\noindent\textbf{Training Details.}
Input image values are normalized to $[-1,1]$. All models are trained using Adam~\cite{Kingma2014AdamAM} optimizer with a base learning rate of 0.002, and a batch size of 8. We perform data augmentation by random horizontal flipping with a probability of 0.5. 

%

\noindent\textbf{Network Architecture.}
Our network architecture is inspired by Pix2PixHD~\cite{wang2018pix2pixHD}. There are four encoder-decoder networks in our pipeline including the flow sub-network, attention mask sub-network, refinement sub-network and attribute removal sub-network. They share similar architectures as U-Net and are composed of three convolutional layers, three residual layers and three deconvolutional layers. For the discriminator we adopt PatchGAN~\cite{Isola2017ImagetoImageTW}, which perform real and fake clasification on local image patches.

\noindent\textbf{Competing Methods.}
We choose state-of-the-art StarGAN~\cite{Choi2017StarGANUG}, UNIT~\cite{Liu2017UnsupervisedIT} and ELEGANT~\cite{Xiao2018ELEGANTEL} as our baselines. StarGAN and UNIT both perform semantic-level facial attribute manipulation, while ELEGANT is capable of transferring instance-level attributes based on given source image.
\subsection{Datasets} 

\noindent\textbf{CelebA}~\cite{liu2015faceattributes} is a large-scale face attributes dataset containing more than 200000 images of celebrity annotated with 40 attributes. We use the standard training, validation and test splits. For preprocessing, we crop images to 178x178, and resize them to 256x256. 

\noindent\textbf{CelebA-HQ}~\cite{Karras2017ProgressiveGO} is a higher quality version of CelebA dataset that allows experimentation with up to 1024x1024 resolution. We only use this dataset as an input to genetate high-resolution results and do not use it for training.

\begin{table*}[t]
\centering
  \begin{tabular}{cc|ccc|ccc|ccc}
  \hline
  \multicolumn{2}{c|}{Metric}                                                                                 & \multicolumn{3}{c|}{FID Score} & \multicolumn{3}{c|}{\begin{tabular}[c]{@{}c@{}}Faithfulness Score\\ \end{tabular}} & \multicolumn{3}{c}{\begin{tabular}[c]{@{}c@{}} Attribute Cls. \\ Accuracy(\%)\end{tabular}} \\ \hline
  \multicolumn{2}{c|}{Attribute}                                                                                 & \rotatebox{90}{Eyeglasses} & \rotatebox{90}{Mustache} & \rotatebox{90}{Goatee} & \rotatebox{90}{Eyeglasses}                        & \rotatebox{90}{Mustache}                         & \rotatebox{90}{Goatee}                           & \rotatebox{90}{Eyeglasses}                      & \rotatebox{90}{Mustache}                      & \rotatebox{90}{Goatee}                     \\ \hline
  \multirow{2}{*}{\begin{tabular}[c]{@{}c@{}}Semantic-level Transfer \end{tabular}} & UNIT                       & \textcolor{red}{0.316}       & \textcolor{red}{0.356}     & \textcolor{red}{0.321}   & \textbackslash{}                  & \textbackslash{}                 & \textbackslash{}                 & 89.5                          & 51.9                          & 61.6                       \\
                                                                                 & StarGAN                    & 0.386       & 0.390     & 0.374   & \textbackslash{}                  & \textbackslash{}                 & \textbackslash{}                 & \textcolor{red}{99.6}                          & \textcolor{red}{98.8}                          & \textcolor{red}{98.8}                       \\ \hline
  \multirow{4}{*}{Instance-level Transfer}                                                & ELEGANT                    & 0.410       & 0.387     & 0.378   & 0.946                      & 0.915                     & 0.905                     & 86.8                          & 82.1                          & 82.3                       \\
                                                                                 & \methodname~(-F) & 0.355       & 0.366     & 0.352   & 0.959                      & 0.883                     & 0.855                     & \textcolor{blue}{96.0}                            & \textcolor{blue}{90.2}                          & 81.5                       \\
                                                                                 & \methodname~(-H)   & 0.353       & 0.337     & 0.324   & \textcolor{blue}{0.786}                      & \textcolor{blue}{0.825}                     & \textcolor{blue}{0.810}                     & 80.7                          & 78.9                          & 77.4                       \\
                                                                                 & \methodname~(full)                 & \textcolor{blue}{0.351}       & \textcolor{blue}{0.336}     & \textcolor{blue}{0.324}   & 0.806                      & 0.832                     & 0.811                     & 91.2                          & 87.8                          & \textcolor{blue}{91.9}                       \\ \hline
  \end{tabular}
  \caption{Benchmarking results of different methods on the CelebA dataset, \wrt three metrics including the FID Score (lower is better), Faithfulness Score (lower is better) and Attribute Classification Accuracy (higher is better), on both semantic-level and instance-level tracks. The red color indicates the best performance in the Semantic-level Transfer track while the blue one indicates the best performance in the Instance-level Transfer track.}
\label{tab:exp-result}
\end{table*}

\subsection{Comparisons to Prior Works}

The comparison is performed on three aspects, including attribute-level face manipulation, instance-level attribute transfer and distribution-level evaluation. We name our approach as \methodname~for easy reference.

\noindent
\textbf{Attribute-level Face Manipulation.}
To evaluate the ability of a method in manipulating a desired attribute, we examined the classification accuracy of synthesized images. We trained three binary facial attribute classifiers for attributes including \textit{Eyeglass, Mustache, Goatee} separately on the CelebA dataset. Using a ResNet-18\cite{He2016DeepRL} architecture, we achieved above $95\%$ accuracy for each of these classifiers. We then trained each of image translation models using the same training set and performed face attribute manipulation on unseen validation set. For ELEGANT and our method, we sampled exemplar images from the same validation set with inverse label. Finally we classified the attribute of these manipulated images using the above-mentioned classifiers. The result is shown in Table~\ref{tab:exp-result}. Though StarGAN achieved the best classification accuracy, the quality of their generated images is poor and all attributes follow the same pattern, as shown in Figure~\ref{fig:compare_mustache}. Our model beats ELEGANT and UNIT at a large margin, suggesting that our model can manipulate attributes more accurately.

\noindent
\textbf{Instance-level Evaluation.}
As StarGAN and UNIT cannot perform attribute transfer at instance level, we compared our method with ELEGANT on this task. To evaluate the faithfulness, we introduced faithfulness score. 

The faithfulness score is designed as follows. Given a target image $a$ and source image $b_y$, the region of desired attribute is cropped as $a^{attr}$ and $b_y^{attr}$ according to facial landmarks. 
We then extract the features of these cropped regions as $f_a$ and $f_{b_y}$ using VGG18~\cite{Simonyan2014VeryDC} pretrained on ImageNet\cite{Deng2009ImageNetAL}. Finally the faithfulness score $s_{faith}$ is computed as the distance of two cropped regions on normalized feature space: 
\begin{equation}
s_{faith}=\norm{\frac{f_a}{\norm{f_a}_2}-\frac{f_{b_y}}{\norm{f_{b_y}}_2}}_2
\end{equation}

Lower score indicates more faithful transfer. As shown in Table~\ref{tab:exp-result}, our method outperforms ELEGANT in term of faithfulness at a large margin. This is because our model introduced flow sub-network to warp pixels directly from target image thus increasing the faithfulness and sharpness of the synthesized image.

\begin{table}
\centering
\begin{tabular}{@{}lll@{}}
\toprule
Method  & \begin{tabular}[c]{@{}l@{}}Attribute\\ Quality\end{tabular} & \begin{tabular}[c]{@{}l@{}}Perceptual \\ Realism\end{tabular} \\ \midrule
StarGAN &              11.3\%                            &           11.9\%                                         \\
UNIT    &           26.1\%                                   &        29.3\%                                             \\
ELEGANT &                  17.4\%                                    &      16.7\%                                                \\
\methodname~(ours)    &              \textbf{45.2\%}                                            &      \textbf{42.1\%}                                              \\ \bottomrule
\end{tabular}
\caption{User study of different methods (the higher the better) \wrt both attribute quality and perceptual realism. Each column represents user preferences that sum to 100\%.}
\label{tab:user-study}
\end{table}

\begin{figure}
  \centering
  \includegraphics[width=\linewidth]{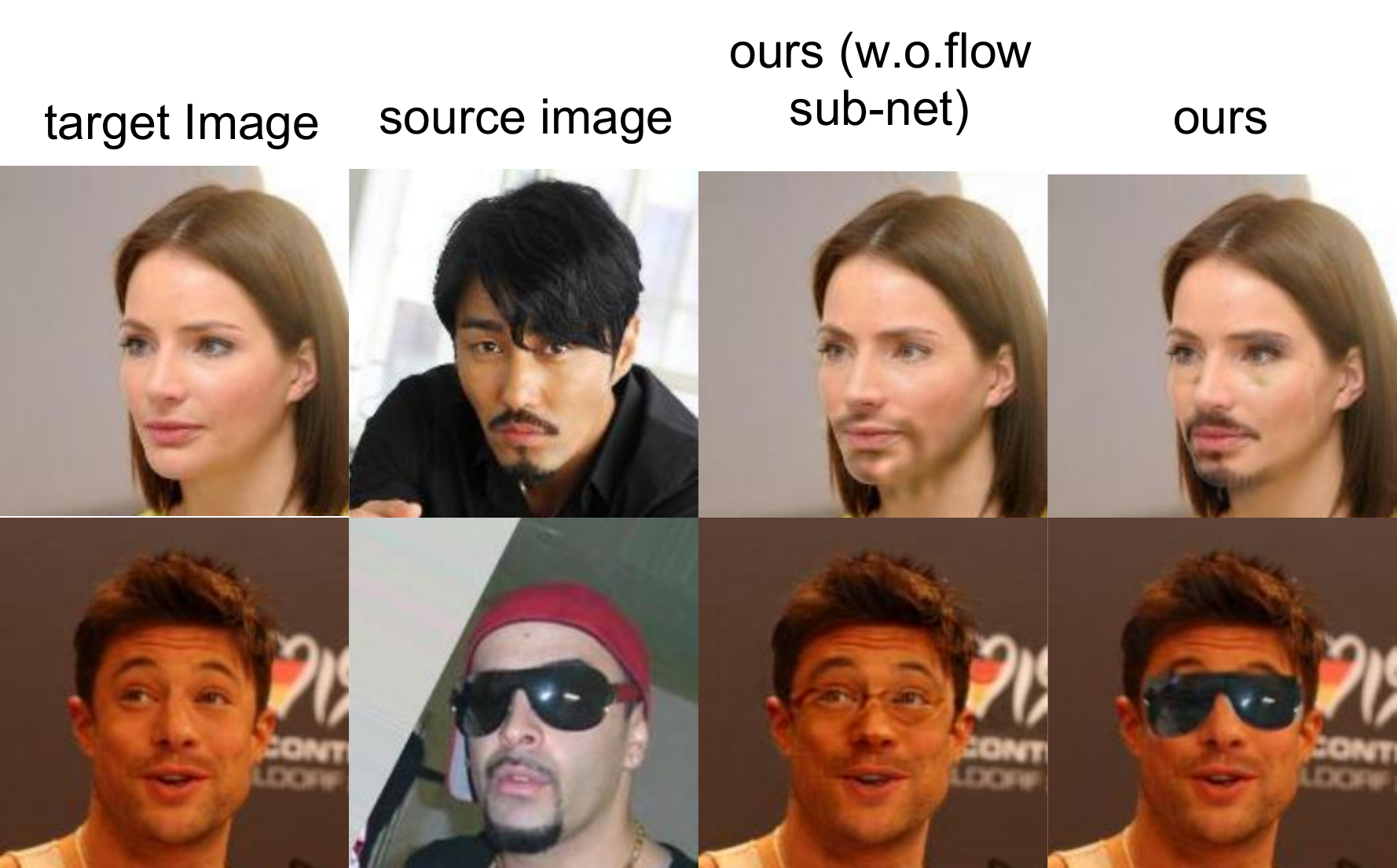}
  \caption{Ablation results of the flow sub-network. From left to right each column represents the target image, source image, our result without flow sub-network and our full result.}
  \label{fig:flowpart_ablation}
\end{figure}

\noindent
\textbf{Distribution-level Evaluation.}
To measure the quality of generated images from different models quantitatively, we calculated Fr\'echet Inception Distance (FID)\cite{Heusel2017GANsTB} between real images and generated images to measure the quality of generated images. The lower the score, the better the result. As shown in Table~\ref{tab:exp-result}, our full model performs better than StarGAN and ELEGANT in all attributes. The FID score of UNIT is relatively low because many generated images look alike the original image and the desired attribute is not manipulated accurately. This is indicated by the low classification accuracy in Table~\ref{tab:exp-result}.

Figure~\ref{fig:compare_mustache} shows examples of images generated when transferring different attributes: the generated images of our model are of high visual quality. These generated images confirm that our model not only preserves the identity of original image, but also captures the attribute to be transferred from source image. Though StarGAN and UNIT have the capability of performing attribute manipulation, many of their results are blurry. Importantly, they can only generate a homogeneous style of attribute for many input images, which lack diversity.

\begin{figure}
  \centering
  \includegraphics[width=\linewidth]{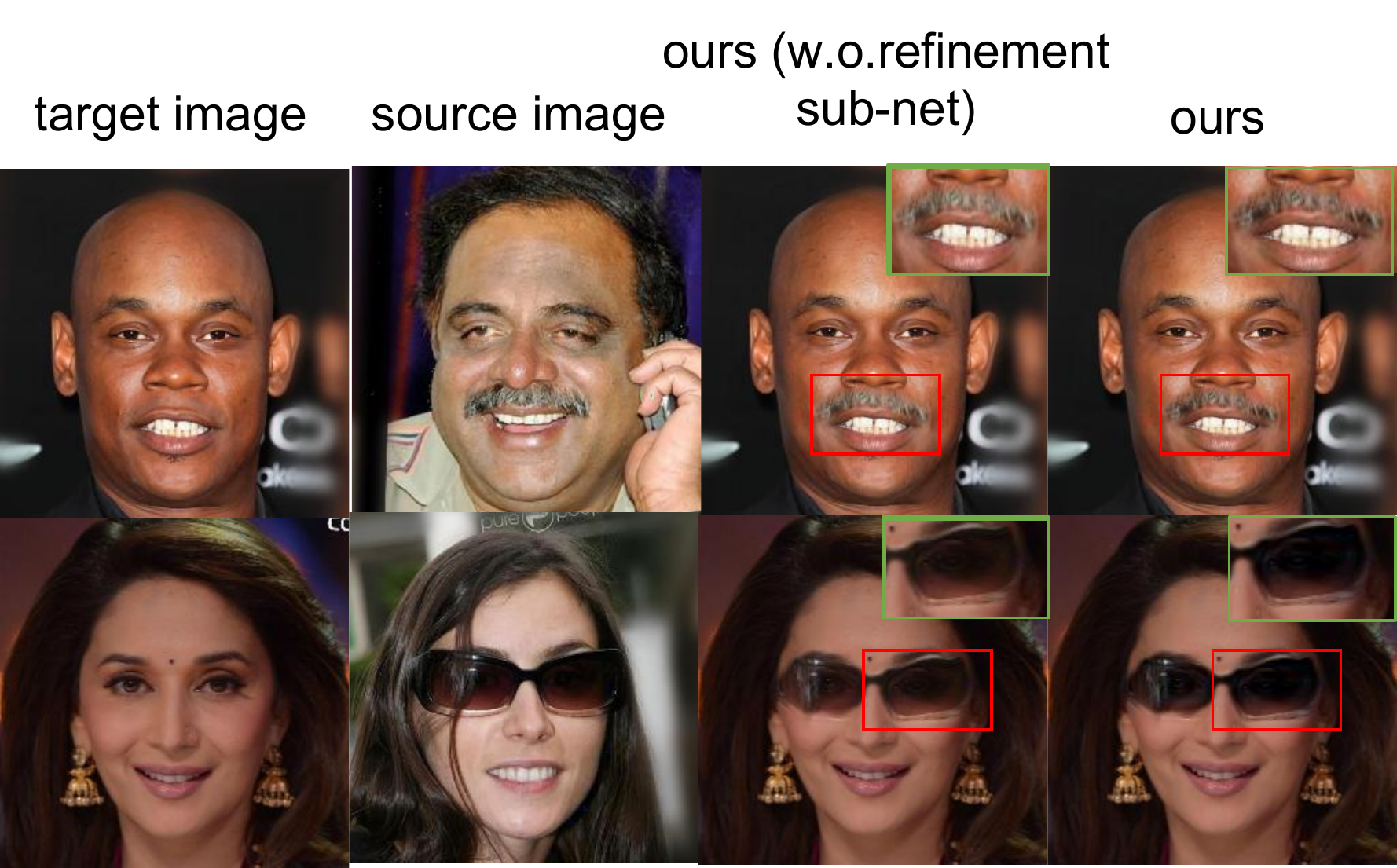}
  \caption{Ablation results of the refinement sub-network. From left to right each column represents target image, source image, our result without refinement sub-network and our full result. }
  \label{fig:hallucination_ablation}
\end{figure}

\noindent
\textbf{User Study.} 
We performed a user survey to assess our model in terms of attribute manipulation and visual quality.  Given an input image, the user were required to choose the best generated image based on two criteria: quality of transfer in attributes and perceptual realism. The options were four randomly shuffled images generated from different methods of the same identity. The modified attribute is selected among \textit{Eyeglasses, Mustache, Goatee} equally. As is shown in Table~\ref{tab:user-study}, our model obtained the majority of votes for the best visual quality while preserving the desired attribute.

\subsection{Ablation Study}

In the ablation study, we considered three variants of our model: (i) \methodname~(full): our full model. (ii) \methodname~(-F): our model with the flow sub-network of attribute transfer module removed, and (iii) \methodname~(-H): our model with the refinement sub-network of attribute transfer module removed. 
Table~\ref{tab:exp-result} lists the FID score, faithfulness score and attribute classification accuracy of these variants. The results suggest the two networks contribute complementary to our full model. More discussion are given below.

\noindent
\textbf{Effectiveness of Geometry-aware Flow.}
Without the flow sub-network, the performance in FID score and faithfulness score dropped significantly, suggesting that the flow sub-network is essential towards faithful facial attribute transfer. Figure~\ref{fig:flowpart_ablation} provides examples of our model without flow sub-network. Without the flow sub-network, synthesized images are blurry and do not preserve features in the source image.

\noindent
\textbf{Effectiveness of Appearance Residual.}
Without the refinement sub-network, a drastic drop of the performance in classification is observed. As shown in Figure~\ref{fig:hallucination_ablation}, without the refinement sub-network the flow sub-network cannot cope with differences in skin colors and lighting conditions.

\begin{figure}
  \centering
  \includegraphics[height=6.5cm]{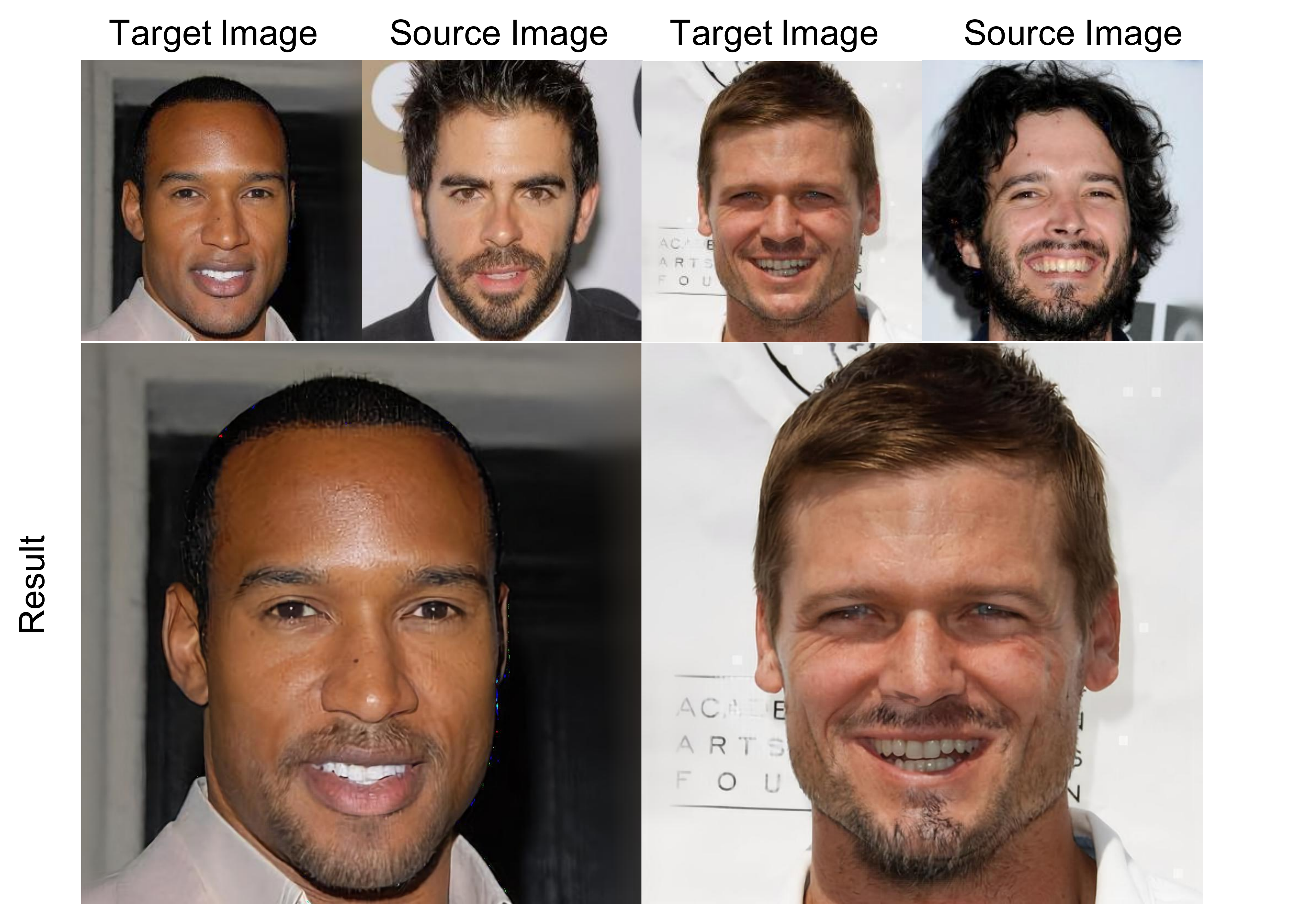}
  \caption{Each triplet includes a source image, target image and our high-res result.}
  \label{fig:high_res_result}
\end{figure}

\subsection{Results on High-Resolution Images}

As the geometry-aware flow is invariant with regard to different resolutions, our model is able to produce high-resolution results without re-training. The result is shown in Figure~\ref{fig:high_res_result}. Given a low-resolution source image and target image, the result is generated by upsampling generated geometry-aware flow, attention mask and appearance residual then apply them on the source image and target image at original resolution. Benefited from the flow sub-network, our model preserves high-frequency patterns in source image and target image well. Scaling other baselines to producing high-resolution results is much harder in comparison to our approach as training on high-resolution images are needed, and the process will be extremely time-consuming and computational ineffective.

\subsection{Results on More Attributes}

\begin{figure}
  \centering
  \includegraphics[width=0.8\linewidth]{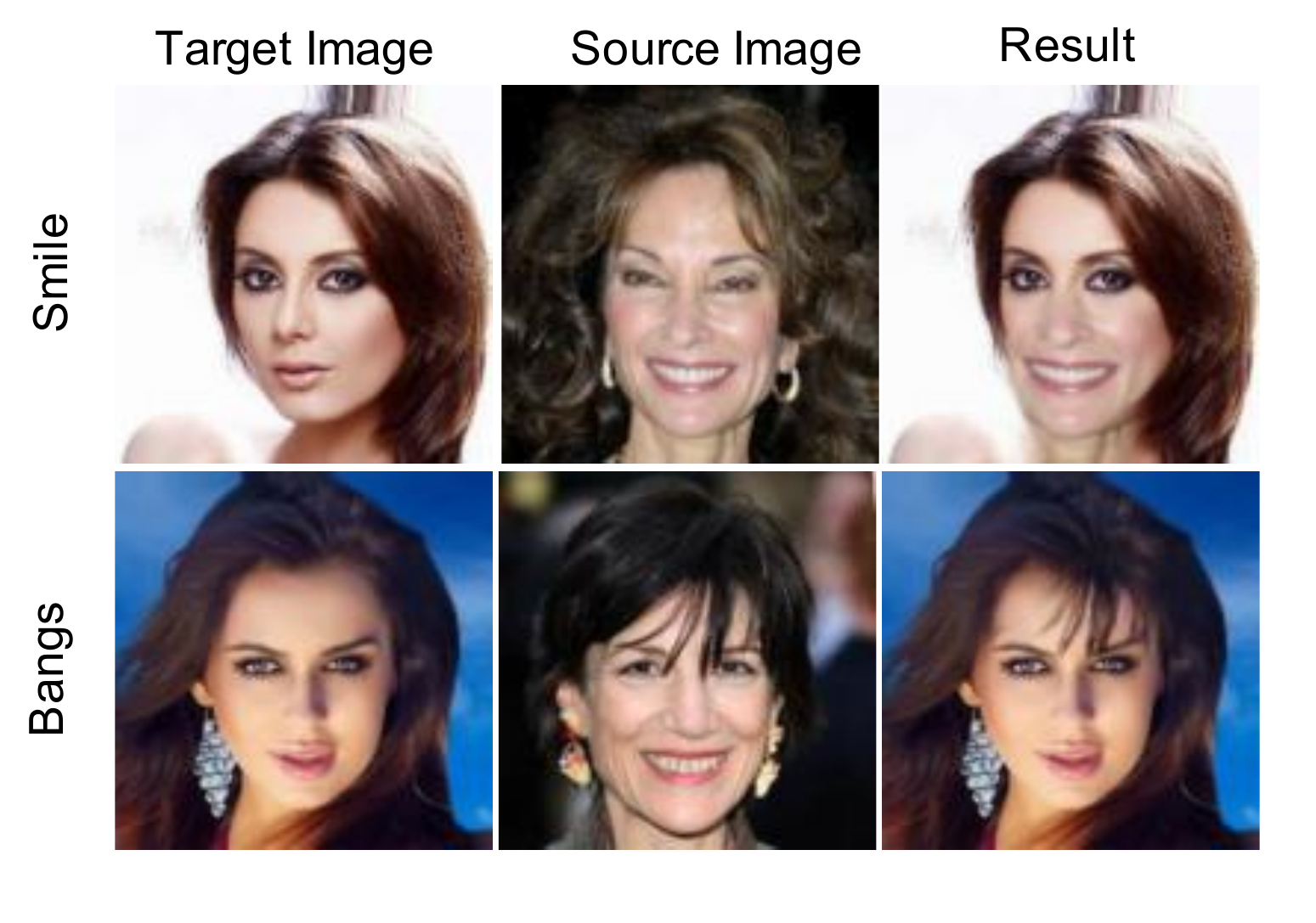}
  \caption{Experiment results on non-local attributes such as smiles and bangs. Every row contains results for one attribute.}
  \label{fig:more_attr}
\end{figure}

As shown in Figure~\ref{fig:more_attr}, our method generalizes well to other challenging attributes such as bangs and smiling which exhibit a high degree of non-local displacement. For “smiling”, the flow subnetwork first warped and blended the teeth and cheek regions into target image, then the refinement subnetwork compensated for the local mis-alignments along the lip. On another non-local attribute “bangs”, our flow network well transferred the bang from the source face to the target face. The global traits such as hair color were addressed by our refinement network.

\section{Conclusion}

We have proposed the notion of geometry-aware flow to address the problem of instance-level facial attribute transfer.
In comparison to prior arts, our method has several appealing properties:
(1) Geometry-aware flow serves as a well-suited representation for modeling the transformation between instance-level facial attributes, despite of the large pose gap between the source and target face images.
(2) When combined with appearance residual produced by the refinement sub-network, our approach is capable of handling potential appearance gap between the source and the target.
(3) Geometry-aware flow can be readily applied to high-resolution face images and generate desired attributes with realistic details.

Though our current framework handles one attribute at a time, it can be readily extended to handle multiple attributes in one pass. For example, following the idea in `StyleBank'~\cite{Chen2017StyleBankAE}, the flow, mask subnetworks and attribute removal network in our framework can be augmented to output multiple flows, masks and image residuals as an attribute bank. During multi-attribute inference, only one element in the `attribute banks' is activated. It is definitely an interesting direction to explore as our future work.

\bibliographystyle{aaai}
\bibliography{library}

\begin{thebibliography}{}

\bibitem[\protect\citeauthoryear{Azadi \bgroup et al\mbox.\egroup
  }{2018}]{Azadi2018CompositionalGL}
Azadi, S.; Pathak, D.; Ebrahimi, S.; and Darrell, T.
\newblock 2018.
\newblock Compositional gan: Learning conditional image composition.
\newblock {\em CoRR} abs/1807.07560.

\bibitem[\protect\citeauthoryear{Brock \bgroup et al\mbox.\egroup
  }{2016}]{Brock2016NeuralPE}
Brock, A.; Lim, T.; Ritchie, J.~M.; and Weston, N.
\newblock 2016.
\newblock Neural photo editing with introspective adversarial networks.
\newblock {\em CoRR} abs/1609.07093.

\bibitem[\protect\citeauthoryear{Bulat and Tzimiropoulos}{2017}]{bulat2017far}
Bulat, A., and Tzimiropoulos, G.
\newblock 2017.
\newblock How far are we from solving the 2d \& 3d face alignment problem? (and
  a dataset of 230,000 3d facial landmarks).
\newblock In {\em ICCV}.

\bibitem[\protect\citeauthoryear{Chen \bgroup et al\mbox.\egroup
  }{2017}]{Chen2017StyleBankAE}
Chen, D.; Yuan, L.; Liao, J.; Yu, N.; and Hua, G.
\newblock 2017.
\newblock Stylebank: An explicit representation for neural image style
  transfer.
\newblock {\em 2017 IEEE Conference on Computer Vision and Pattern Recognition
  (CVPR)}  2770--2779.

\bibitem[\protect\citeauthoryear{Choi \bgroup et al\mbox.\egroup
  }{2017}]{Choi2017StarGANUG}
Choi, Y.; Choi, M.-J.; Kim, M.; Ha, J.-W.; Kim, S.; and Choo, J.
\newblock 2017.
\newblock Stargan: Unified generative adversarial networks for multi-domain
  image-to-image translation.
\newblock {\em CoRR} abs/1711.09020.

\bibitem[\protect\citeauthoryear{Deng \bgroup et al\mbox.\egroup
  }{2009}]{Deng2009ImageNetAL}
Deng, J.; Dong, W.; Socher, R.; Li, L.-J.; Li, K.; and Fei-Fei, L.
\newblock 2009.
\newblock Imagenet: A large-scale hierarchical image database.
\newblock {\em CVPR}.

\bibitem[\protect\citeauthoryear{Gardner \bgroup et al\mbox.\egroup
  }{2015}]{Gardner2015DeepMT}
Gardner, J.~R.; Kusner, M.~J.; Li, Y.; Upchurch, P.; Weinberger, K.~Q.; and
  Hopcroft, J.~E.
\newblock 2015.
\newblock Deep manifold traversal: Changing labels with convolutional features.
\newblock {\em CoRR} abs/1511.06421.

\bibitem[\protect\citeauthoryear{He \bgroup et al\mbox.\egroup
  }{2016}]{He2016DeepRL}
He, K.; Zhang, X.; Ren, S.; and Sun, J.
\newblock 2016.
\newblock Deep residual learning for image recognition.
\newblock {\em CVPR}.

\bibitem[\protect\citeauthoryear{Heusel \bgroup et al\mbox.\egroup
  }{2017}]{Heusel2017GANsTB}
Heusel, M.; Ramsauer, H.; Unterthiner, T.; Nessler, B.; and Hochreiter, S.
\newblock 2017.
\newblock Gans trained by a two time-scale update rule converge to a local nash
  equilibrium.
\newblock In {\em NIPS}.

\bibitem[\protect\citeauthoryear{Huang \bgroup et al\mbox.\egroup
  }{2018}]{huang2018deep}
Huang, C.; Li, Y.; Loy, C.~C.; and Tang, X.
\newblock 2018.
\newblock Deep imbalanced learning for face recognition and attribute
  prediction.
\newblock {\em arXiv preprint arXiv:1806.00194}.

\bibitem[\protect\citeauthoryear{Isola \bgroup et al\mbox.\egroup
  }{2017}]{Isola2017ImagetoImageTW}
Isola, P.; Zhu, J.-Y.; Zhou, T.; and Efros, A.~A.
\newblock 2017.
\newblock Image-to-image translation with conditional adversarial networks.
\newblock {\em CVPR}.

\bibitem[\protect\citeauthoryear{Jaderberg \bgroup et al\mbox.\egroup
  }{2015}]{Jaderberg2015SpatialTN}
Jaderberg, M.; Simonyan, K.; Zisserman, A.; and Kavukcuoglu, K.
\newblock 2015.
\newblock Spatial transformer networks.
\newblock In {\em NIPS}.

\bibitem[\protect\citeauthoryear{Karras \bgroup et al\mbox.\egroup
  }{2017}]{Karras2017ProgressiveGO}
Karras, T.; Aila, T.; Laine, S.; and Lehtinen, J.
\newblock 2017.
\newblock Progressive growing of gans for improved quality, stability, and
  variation.
\newblock {\em CoRR} abs/1710.10196.

\bibitem[\protect\citeauthoryear{Kingma and Ba}{2014}]{Kingma2014AdamAM}
Kingma, D.~P., and Ba, J.
\newblock 2014.
\newblock Adam: A method for stochastic optimization.
\newblock {\em CoRR} abs/1412.6980.

\bibitem[\protect\citeauthoryear{Kingma and
  Welling}{2013}]{Kingma2013AutoEncodingVB}
Kingma, D.~P., and Welling, M.
\newblock 2013.
\newblock Auto-encoding variational bayes.
\newblock {\em CoRR} abs/1312.6114.

\bibitem[\protect\citeauthoryear{Lample \bgroup et al\mbox.\egroup
  }{2017}]{Lample2017FaderNM}
Lample, G.; Zeghidour, N.; Usunier, N.; Bordes, A.; Denoyer, L.; and Ranzato,
  M.
\newblock 2017.
\newblock Fader networks: Manipulating images by sliding attributes.
\newblock In {\em NIPS}.

\bibitem[\protect\citeauthoryear{Lin \bgroup et al\mbox.\egroup
  }{2018}]{Lin2018STGANST}
Lin, C.-H.; Yumer, E.; Wang, O.; Shechtman, E.; and Lucey, S.
\newblock 2018.
\newblock St-gan: Spatial transformer generative adversarial networks for image
  compositing.
\newblock {\em CoRR} abs/1803.01837.

\bibitem[\protect\citeauthoryear{Liu and Tuzel}{2016}]{Liu2016CoupledGA}
Liu, M.-Y., and Tuzel, O.
\newblock 2016.
\newblock Coupled generative adversarial networks.
\newblock In {\em NIPS}.

\bibitem[\protect\citeauthoryear{Liu \bgroup et al\mbox.\egroup
  }{2015}]{liu2015faceattributes}
Liu, Z.; Luo, P.; Wang, X.; and Tang, X.
\newblock 2015.
\newblock Deep learning face attributes in the wild.
\newblock In {\em ICCV}.

\bibitem[\protect\citeauthoryear{Liu \bgroup et al\mbox.\egroup
  }{2017}]{liu2017video}
Liu, Z.; Yeh, R.~A.; Tang, X.; Liu, Y.; and Agarwala, A.
\newblock 2017.
\newblock Video frame synthesis using deep voxel flow.
\newblock In {\em ICCV}.

\bibitem[\protect\citeauthoryear{Liu, Breuel, and
  Kautz}{2017}]{Liu2017UnsupervisedIT}
Liu, M.-Y.; Breuel, T.; and Kautz, J.
\newblock 2017.
\newblock Unsupervised image-to-image translation networks.
\newblock In {\em NIPS}.

\bibitem[\protect\citeauthoryear{Loy, Luo, and Huang}{2017}]{loy2017deep}
Loy, C.~C.; Luo, P.; and Huang, C.
\newblock 2017.
\newblock Deep learning face attributes for detection and alignment.
\newblock In {\em Visual Attributes}. Springer.
\newblock  181--214.

\bibitem[\protect\citeauthoryear{Mao \bgroup et al\mbox.\egroup
  }{2017}]{Mao2017LeastSG}
Mao, X.; Li, Q.; Xie, H.; Lau, R. Y.~K.; Wang, Z.; and Smolley, S.~P.
\newblock 2017.
\newblock Least squares generative adversarial networks.
\newblock {\em 2017 IEEE International Conference on Computer Vision (ICCV)}
  2813--2821.

\bibitem[\protect\citeauthoryear{Nguyen \bgroup et al\mbox.\egroup
  }{2008}]{Nguyen2008ImagebasedS}
Nguyen, M.~H.; Lalonde, J.-F.; Efros, A.~A.; and la~Torre, F.~D.
\newblock 2008.
\newblock Image-based shaving.
\newblock {\em Computer Graphics Forum}.

\bibitem[\protect\citeauthoryear{Perarnau \bgroup et al\mbox.\egroup
  }{2016}]{Perarnau2016InvertibleCG}
Perarnau, G.; van~de Weijer, J.; Raducanu, B.; and {\'A}lvarez, J.~M.
\newblock 2016.
\newblock Invertible conditional gans for image editing.
\newblock {\em CoRR} abs/1611.06355.

\bibitem[\protect\citeauthoryear{Shen and Liu}{2017}]{Shen2017LearningRI}
Shen, W., and Liu, R.
\newblock 2017.
\newblock Learning residual images for face attribute manipulation.
\newblock {\em CVPR}.

\bibitem[\protect\citeauthoryear{Shu \bgroup et al\mbox.\egroup
  }{2017}]{Shu2017NeuralFE}
Shu, Z.; Yumer, E.; Hadap, S.; Sunkavalli, K.; Shechtman, E.; and Samaras, D.
\newblock 2017.
\newblock Neural face editing with intrinsic image disentangling.
\newblock {\em 2017 IEEE Conference on Computer Vision and Pattern Recognition
  (CVPR)}  5444--5453.

\bibitem[\protect\citeauthoryear{Simonyan and
  Zisserman}{2014}]{Simonyan2014VeryDC}
Simonyan, K., and Zisserman, A.
\newblock 2014.
\newblock Very deep convolutional networks for large-scale image recognition.
\newblock {\em CoRR} abs/1409.1556.

\bibitem[\protect\citeauthoryear{Wang \bgroup et al\mbox.\egroup
  }{2018}]{wang2018pix2pixHD}
Wang, T.-C.; Liu, M.-Y.; Zhu, J.-Y.; Tao, A.; Kautz, J.; and Catanzaro, B.
\newblock 2018.
\newblock High-resolution image synthesis and semantic manipulation with
  conditional gans.
\newblock In {\em Proceedings of the IEEE Conference on Computer Vision and
  Pattern Recognition}.

\bibitem[\protect\citeauthoryear{Xiao, Hong, and Ma}{2018}]{Xiao2018ELEGANTEL}
Xiao, T.; Hong, J.; and Ma, J.
\newblock 2018.
\newblock Elegant: Exchanging latent encodings with gan for transferring
  multiple face attributes.
\newblock {\em CoRR} abs/1803.10562.

\bibitem[\protect\citeauthoryear{Yeh \bgroup et al\mbox.\egroup
  }{2016}]{Yeh2016SemanticFE}
Yeh, R.~A.; Liu, Z.; Goldman, D.~B.; and Agarwala, A.
\newblock 2016.
\newblock Semantic facial expression editing using autoencoded flow.
\newblock {\em CoRR} abs/1611.09961.

\bibitem[\protect\citeauthoryear{Zhou \bgroup et al\mbox.\egroup
  }{2016}]{Zhou2016ViewSB}
Zhou, T.; Tulsiani, S.; Sun, W.; Malik, J.; and Efros, A.~A.
\newblock 2016.
\newblock View synthesis by appearance flow.
\newblock In {\em ECCV}.

\bibitem[\protect\citeauthoryear{Zhou \bgroup et al\mbox.\egroup
  }{2017}]{Zhou2017GeneGANLO}
Zhou, S.; Xiao, T.; Yang, Y.; Feng, D.; He, Q.; and He, W.
\newblock 2017.
\newblock Genegan: Learning object transfiguration and attribute subspace from
  unpaired data.
\newblock {\em CoRR} abs/1705.04932.

\end{thebibliography}
\clearpage
\onecolumn
\section{Supplementary Materials}
\subsection{Additional Quanlitative Results on CelebA dataset}
\label{sec:supp_celeba}
\begin{figure}[!htb]
  \centering
  \includegraphics[width=\linewidth]{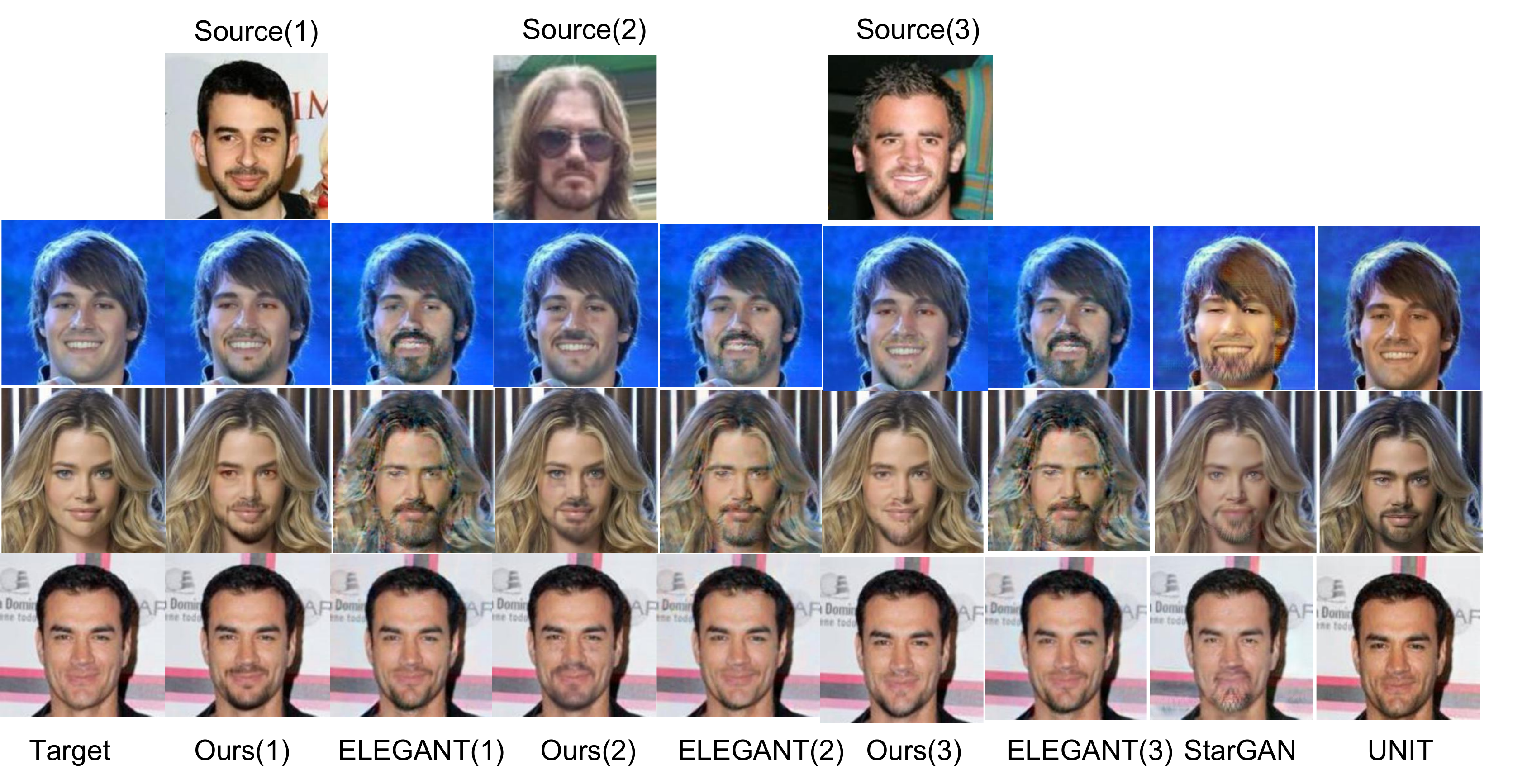}
  \caption{Goatee transfer results on CelebA dataset. }
\end{figure}
\begin{figure}[!htb]
  \centering
  \includegraphics[width=\linewidth]{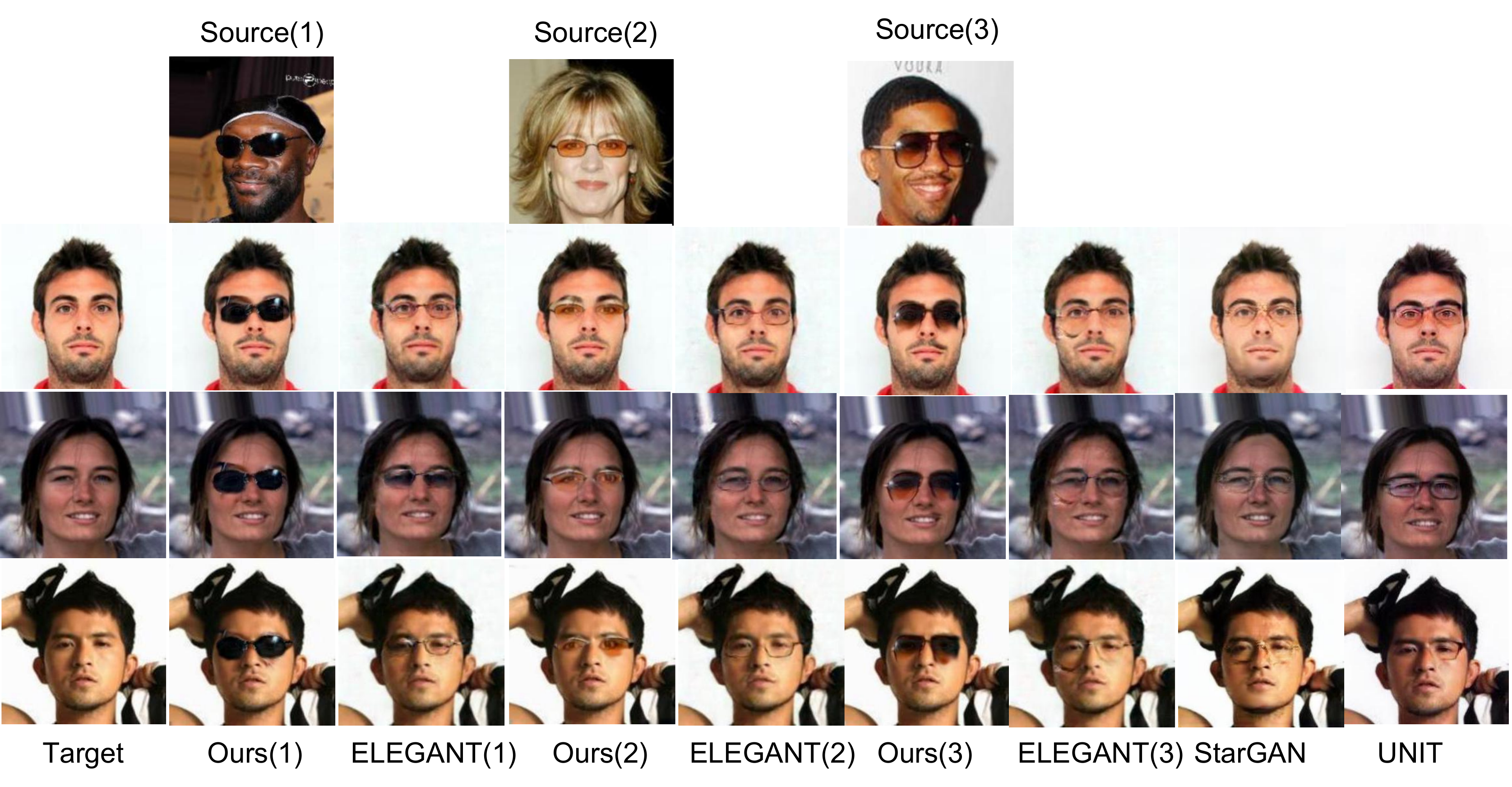}
  \caption{Eyeglass transfer results on CelebA dataset. }
\end{figure}
\subsection{Additional Quanlitative Results on CelebA-HQ dataset}
\label{sec:supp_celeba_hq}
\begin{figure}[!htb]
  \centering
  \includegraphics[width=\linewidth]{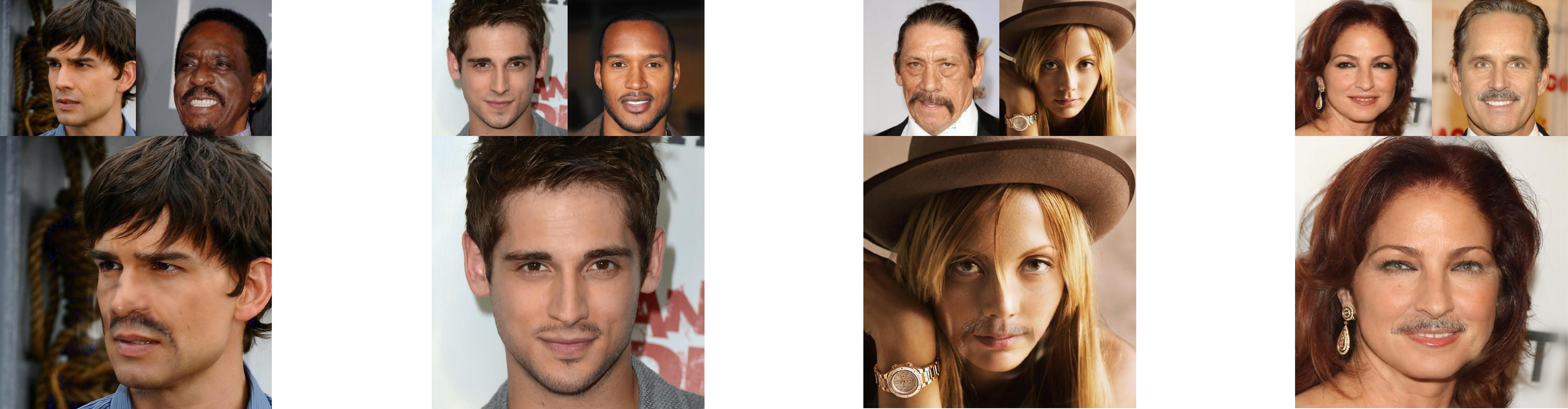}
  \caption{Mustache transfer results on CelebA-HQ dataset. }
\end{figure}
\begin{figure}[!htb]
  \centering
  \includegraphics[width=\linewidth]{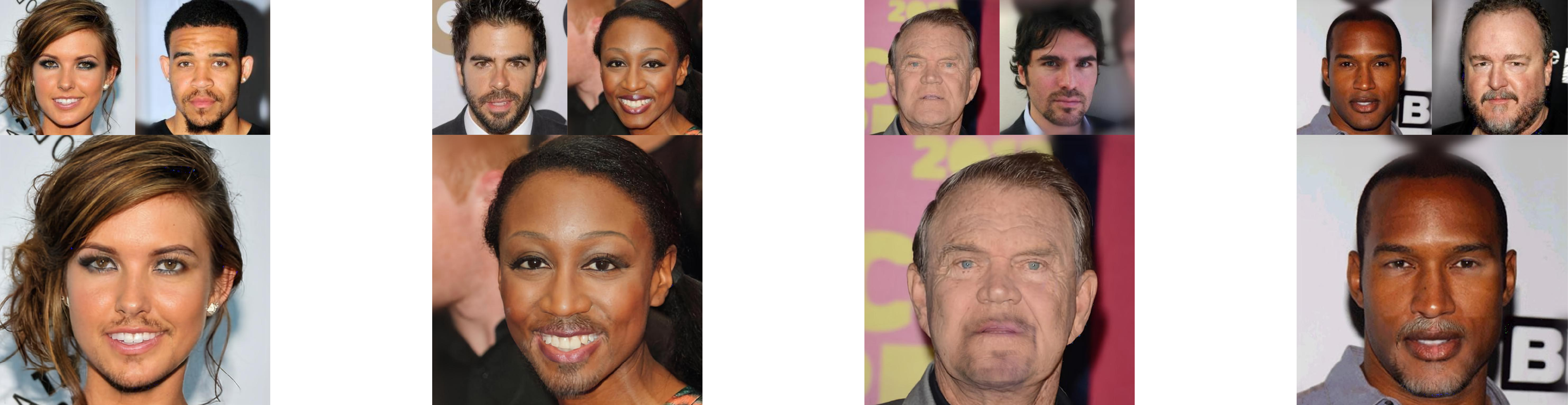}
  \caption{Goatee transfer results on CelebA-HQ dataset. }
\end{figure}
\begin{figure}[!htb]
  \centering
  \includegraphics[width=\linewidth]{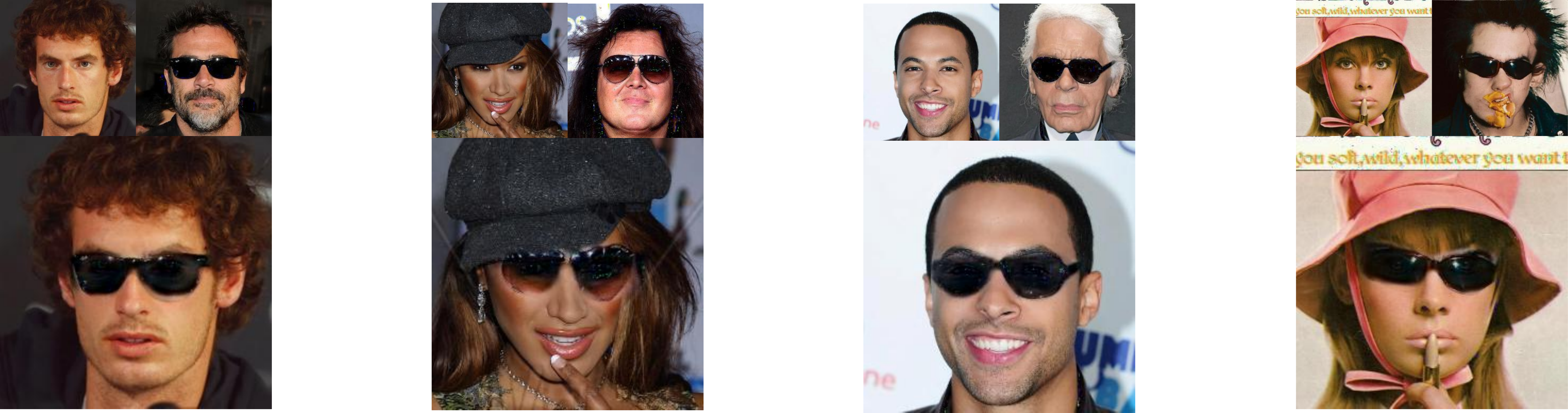}
  \caption{Eyeglass transfer results on CelebA-HQ dataset. }
\end{figure}
\end{document}